\documentclass[conference]{IEEEtran}
\IEEEoverridecommandlockouts
\usepackage{amsmath,amssymb,amsfonts}
\usepackage{algorithmic}
\usepackage{graphicx}
\usepackage{textcomp}
\usepackage{xcolor}
\usepackage{multirow}
\usepackage{adjustbox}
\usepackage[
backend=biber,
style=alphabetic,
sorting=ynt
]{biblatex}
\usepackage{subcaption} 
\def\BibTeX{{\rm B\kern-.05em{\sc i\kern-.025em b}\kern-.08em
    T\kern-.1667em\lower.7ex\hbox{E}\kern-.125emX}}
\begin{document}

\title{Combating Digitally Altered Images: Deepfake Detection\\}

\author{\IEEEauthorblockN{Saksham Kumar}
\IEEEauthorblockA{\textit{Amrita School of Computing} \\
\textit{Amrita Vishwa Vidyapeetham}\\
Amaravati(Andhra Pradesh), India \\
sakshamkumarprasad03@gmail.com}
\and
\IEEEauthorblockN{Rhythm Narang}
\IEEEauthorblockA{\textit{Dept of Computer Science and Engineering} \\
{\textit{ Thapar Institute of Engineering \& Technology}} 
\\ Patiala(Punjab), India \\
narangrhythm.nr@gmail.com}
}
\maketitle

\begin{abstract}
The rise of Deepfake technology to generate hyper-realistic manipulated images and videos poses a significant challenge to the public and relevant authorities. This study presents a robust Deepfake detection based on a modified Vision Transformer(ViT) model, trained to distinguish between real and Deepfake images. The model has been trained on a subset of the OpenForensics Dataset with multiple augmentation techniques to increase robustness for diverse image manipulations. The class imbalance issues are handled by oversampling and a train-validation split of the dataset in a stratified manner. Performance is evaluated using the accuracy metric on the training and testing datasets, followed by a prediction score on a random image of people, irrespective of their realness. The model demonstrates state-of-the-art results on the test dataset to meticulously detect Deepfake images. 
\end{abstract}


\section{Introduction}
In recent years, the capabilities of artificial intelligence (AI) have advanced rapidly, enabling the creation of hyper-realistic synthetic media, commonly referred to as "Deepfake" images. Deepfake leverage techniques in deep learning and computer graphics to generate or alter video, audio, and image content, often rendering it indistinguishable from real media to the human eye. While these technologies have practical applications in fields such as entertainment and education, their misuse has led to widespread societal concerns. Weaponized Deepfake pose serious risks to privacy, security, and the well-being of individuals, particularly vulnerable populations, by enabling the spread of misinformation and personal defamation, which undermines trust in digital content. \cite{electronics13010095}\\\
\newline
Recent advancements in digital technology and AI have significantly enhanced the ability to create highly realistic images and videos, making it increasingly challenging to differentiate between genuine and fabricated media. This capability, often harnessed through "Deepfake" technology—derived from the terms "deep learning" and "fake"—enables the manipulation of faces, voices, and expressions in images and videos, often undetectable to the human eye. While Deepfake technology has legitimate applications, it also poses serious privacy and security risks, raising concerns about misuse.\cite{10.1007/978-981-16-0733-2_39}\\\

Deep Learning (DL) has been widely applied in healthcare, industry, and academia to tackle complex challenges, such as medical diagnosis and large data analytics. However, advancements in digital technology have also led to the rise of applications like Deepfake, which pose significant threats to democracy, national security, and privacy. Deepfake systems use DL techniques to create highly convincing fake media—images, videos, and audio—that are virtually indistinguishable from real content. This capability has raised concerns about misinformation, fraud, and misuse in legal settings. To address these issues, it is essential to develop technologies that can detect such falsified content. This study provides a comprehensive review of Deepfake detection methods, categorizing them into video, image, audio, and hybrid multimedia detection approaches.\\

This research focuses on the classification of Deepfake images using a Vision Transformer (ViT) model. The model is trained and evaluated on the OpenForensics dataset, which comprises a diverse set of Deepfake and real images. Below, we describe the architectural components and preprocessing steps involved in building our classification model.\\

\begin{table*}[ht]
\centering
\scriptsize
\renewcommand{\arraystretch}{1.2}
\begin{adjustbox}{max width=\textwidth}
\begin{tabular}{|p{3.5cm}|p{4.5cm}|p{3.5cm}|p{4.5cm}|}
\hline
\textbf{Study} & \textbf{Method} & \textbf{Features} & \textbf{Key Findings} \\
\hline
\textbf{CNN-based Method \cite{LIN2023103895}} & Multi-scale convolution with dilation and depthwise separable convolution, plus vision transformer for global features. & Multi-scale module captures face details and tampering artifacts. & Achieves strong detection on both high- and low-quality datasets; good cross-dataset generalization. Experiments confirm better performance than related methods; ablation studies validate model components. \\
\hline
\textbf{DEEPFAKER1 \cite{10.1145/3634914}} & Unified platform integrating 10 deepfake and 9 detection methods. & Modular, user-friendly design for large-scale testing. & Detection methods show weak generalization across deepfake types; APIs are vulnerable to adversarial deepfakes with a 70\% success rate. Lab methods are more robust than real-world APIs; video compression does not always increase detection difficulty. \\
\hline
\textbf{Generative Algorithms \cite{TOLOSANA2020131}} & Deepfake generation via deep learning algorithms focused on face manipulation. & Overview of current algorithms and datasets used in deepfake creation. & Highlights the challenges of detecting deepfakes due to realistic outputs created by accessible applications. Discusses both the beneficial and unethical uses of deepfake technology in real-world applications. \\
\hline
\end{tabular}
\end{adjustbox}
\caption{Literature Review - Deepfake Detection }
\end{table*}
\subsection{Literature Review}

The accessibility of advanced multimedia manipulation tools like FaceSwap and deepfake technologies has enabled realistic synthetic media creation, raising concerns over misinformation and defamation in digital content. Various studies have explored deepfake detection using machine learning (ML) and deep learning (DL), particularly generative adversarial networks (GANs), to address these issues. Verdoliva et al.\cite{Verdoliva_2020} offers a comprehensive overview of traditional and DL-based visual media verification methods, while Tolosana et al. \cite{TOLOSANA2020131} categorize deepfake manipulations into types such as identity swaps and expression modifications, analyzing the challenges and future trends in detection technologies.

Mirsky and Lee \cite{Mirsky_2021} emphasize the mechanics and detection techniques of human reenactment deepfakes, highlighting gaps in current defenses. Castillo Camacho and Wang \cite{jimaging7040069} focus on image forensics, categorizing DL-based detection methods and addressing key issues in deepfake image identification. Similarly, Yu et al. \cite{https://doi.org/10.1049/bme2.12031} discuss video deepfake detection, identifying generalization challenges for robust detection models, while Rana et al. \cite{9721302} compare DL, traditional ML, statistical, and blockchain-based methods, concluding that DL approaches are the most effective in detecting deepfakes.

Despite this progress, current literature is limited in scope, often lacking comprehensive comparisons across deepfake types and media formats. This study aims to bridge these gaps by reviewing DL-based detection techniques for diverse media, including images and hybrid content, thus providing a more extensive analysis of available methods and datasets for advancing detection capabilities.

\section{Methodology}
The model architecture used was based on the Vision Transformer (ViT), known for its effectiveness in various image classification tasks due to its transformer-based architecture, which captures global relationships within the image.

\subsection{Dataset}
For this study, we utilized the OpenForensics Dataset, a well-known dataset in deepfake detection research. The dataset contains both real and synthetically generated fake images through advanced manipulation techniques. The dataset was divided into training, validation, and testing sets, in ratio 14:4:1, respectively.

\subsection{Preprocessing and Data Augmentation}
Data preprocessing was applied to enhance the robustness of our deepfake classification model. Specifically, we applied image transformations compatible with the input requirements of a Vision Transformer (ViT) model, particularly the "google vit-base-patch16-224-in21k" model. This model is pre-trained on a large collection of images in a supervised fashion, namely ImageNet-21k, at a resolution of \(224 \times 224\) pixels.

\subsection{Model Architecture}

In the proposed work, we utilise a modified version of Vision Transformers(ViT) \cite{dosovitskiy2021imageworth16x16words}, i.e., Google ViT-base-patch16-224-in21k \cite{deng2009imagenet} pre-trained model. The ViT model leverages a transformer architecture originally developed for text but applied to images . The consecutive steps:
\newline
\begin{itemize}
    \item Dividing image into patches: Each input image of size \(s \times s\) is divided into smaller patches of size \(p\times p\), making \(N = \frac{s^2}{p^2}\) patches. Each patch is flattened, followed by linear projection for vector representation.\\\
    \item Positional with Transformer Encoder: Positional encodings were added to patch embeddings to retain spatial information. The patches are then fed into the transformer encoder, comprising of multi-head self-attention layers and feed-forward neural networks.\\\
    \item Classification Output: A fully connected layer yields the classification probability for both real or fake class labels.\\\
\end{itemize}
\begin{figure}
    \centering
    \includegraphics[width=0.9\linewidth]{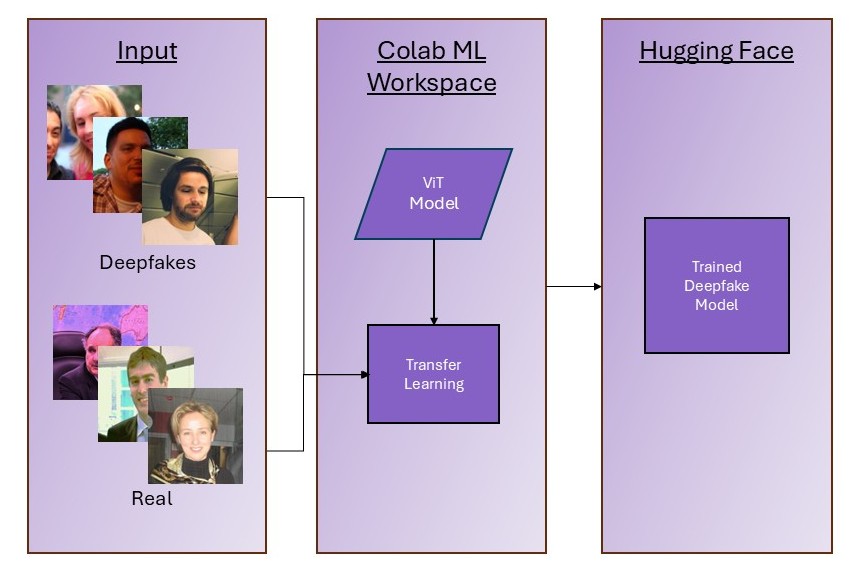}
    \caption{ViT Model Architecture}
    \label{fig:enter-label}
\end{figure}
\begin{figure*}[htp]
    \centering
    \begin{subfigure}[b]{0.2\textwidth}
        \centering
        \includegraphics[height=0.2\textheight]{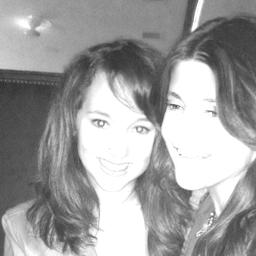}
        \caption{Original Label: Real \\ Real Prob: 0.9807 \\ Fake Prob: 0.0225 \\ Predicted Label: Real}
    \end{subfigure}
    \hfill
    \begin{subfigure}[b]{0.2\textwidth}
        \centering
        \includegraphics[height=0.2\textheight]{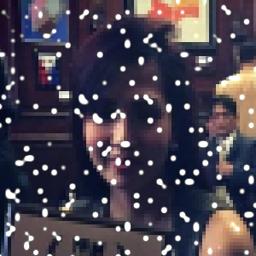}
        \caption{Original Label: Real \\ Real Prob: 0.9115 \\ Fake Prob: 0.0885 \\ Predicted Label: Real}
    \end{subfigure}
    \hfill
    \begin{subfigure}[b]{0.2\textwidth}
        \centering
        \includegraphics[height=0.2\textheight]{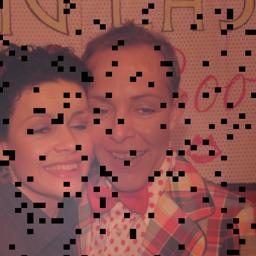}
        \caption{Original Label: Fake \\ Real Prob: 0.1253 \\ Fake Prob: 0.8747 \\ Predicted Label: Fake}
    \end{subfigure}
    \hfill
    \begin{subfigure}[b]{0.2\textwidth}
        \centering
        \includegraphics[height=0.2\textheight]{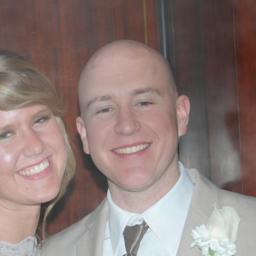}
        \caption{Original Label: Fake \\ Real Prob: 0.0956 \\ Fake Prob: 0.9044 \\ Predicted Label: Fake}
    \end{subfigure}
    
    \caption{Classification results for real and fake images, showing original label, probability scores, and predicted label.}
    \label{fig:results}
\end{figure*}
\subsection{Training}
The model was fine-tuned on the OpenForensics dataset using the Adam optimizer with a learning rate of \(\eta\). The training process was computed with categorical cross-entropy loss between the predicted probabilities and ground truth labels. The categorical cross-entropy loss function can be expressed as:

\[L = - \sum_{i=1}^{C} y_i \cdot \log(\hat{y}_i)\]
where L represents the loss function, C is the number of classes, \(y_i\) is the true label (1 for correct, 0 otherwise) and \(\hat{y}_i\) is the predicted probability for class i.\\\

\section{Results}
The results of our modified ViT model for Deepfake Image Classification on the OpenForensics dataset showcase the model efficiency in accurately classifying real and fake images. We have limited the number of epochs to 2, given the extent and resource-intensive nature of the model and quality of ViT model even under minimal training. We evaluate the training and validation results over two epochs. 
\subsection{Model Runtime and Efficiency}
The model’s evaluation runtime of approximately 799.47 seconds and 95.264 samples per second is suggestive of optimal efficiency, processing a high number of images per second.
\subsection{Training and Validation}
\begin{table}
    \centering
\caption{Training and Validation Loss}
\label{Training and Validation Loss}
    \begin{tabular}{|c|c|c|} \hline 
         Epoch&  Training Loss& Validation Loss\\ \hline 
         1&  0.0515& 0.023838\\ \hline 
         2&  0.045& 0.023709\\ \hline
    \end{tabular}    
\end{table}

\begin{itemize}
    \item The gradual decrease training loss from 0.0515 to 0.0415 indicates effective model training on the provided training data.
    \item The validation loss remains almost remains constant to an accuracy of 4 significant figures 0.023838 to 0.023709 indicate the model's stabilisation and optimality.
\end{itemize}

\subsection{Accuracy}
The model produces state-of-the art result over the testing dataset \( \geqslant 99\% \), the results being discussed below: 
\begin{table}
    \centering
\caption{Evaluation Accuracy}
\label{Evaluation Accuracy}
    \begin{tabular}{|c|c|} \hline 
         Metric& Value\\ \hline 
         Evaluation Accuracy& 0.992043171\\ \hline 
         Evaluation Loss& 0.023709\\ \hline 
         Samples/sec& 95.264\\ \hline 
         Steps per second& 5.955\\ \hline
    \end{tabular}

\end{table}

\subsection{Inference on Real and Fake Images}
Inferences on multiple Real and Deepfake images were given. The images taken had multiple real world issues like blurriness, over and under exposure, multiple angles and pixel loss, yet the model is able to provide paramount results.
\begin{itemize}
    \item Real Images would achieve a greater probability on real than fake with value of real nearing 1. 
    \item Fake Images would achieve a greater probability on fake than real with value of fake nearing 1. 
\end{itemize}
\section{Conclusion}
The model’s results demonstrate not only high accuracy but also effective learning with minimal validation loss, suggesting that it is well-optimized for the task of deepfake detection. This performance, paired with efficient processing, positions the model for practical use in real-world applications. To further enhance the model’s performance, additional fine-tuning, more diverse datasets and more epochs could be explored to ensure sustained improvement, particularly in edge cases or for harder-to-detect deepfakes.

\printbibliography
\end{document}